RESEARCH ARTICLE

# Using GeoAI and machine learning tools for consistent high-resolution land cover mapping based on time-series NAIP imagery

**Jie Liu · Xusheng Tang · Chao Wang · Zhengxiao Yan · Yuchen Dai · Qi Zhang · Conghe Song**

**Abstract**
*Context* High-resolution land-cover maps are essential for ecological monitoring and landscape-level analysis, yet long-term high-resolution time-series products remain scarce. The National Agricultural Imagery Program (NAIP) provides highly valuable aerial imagery, but strong temporal heterogeneity across years, primarily caused by shifts in sensor characteristics, limits the ability to generate consistent multi-year land cover maps.
*Objectives* This study aims to (1) develop an algorithm capable of producing spatially detailed and temporally coherent 1-m land-cover maps using the state-of-the-art GeoAI/Machine Learning (ML) tools based on time-series NAIP imagery, and (2) address cross-year sensor characteristic shifts without relying on historical training samples.
*Methods* We designed a dual-track adaptive workflow that applies different strategies to NAIP imagery with different qualities. NAIP imagery collected during 2009–2017 has a higher quality than that collected during 2004–2008. Images from the high-quality years were classified using a foundation model pretrained with U-Net/ResNet-34 and refined with a Segment Anything Model (SAM) for accurate boundary delineation. Images collected in the earlier years were reconstructed using an NLCD-based spatiotemporal bridging and label back-casting pipeline. Accuracy was evaluated across six U.S. counties in North Carolina and Pennsylvania.
*Results* The algorithm produced stable results across years, yielding overall accuracies of 0.887 (2014), 0.886 (2017), and 0.733 (2004) with Kappa statistics at 0.860, 0.831, and 0.764, respectively. Structure, Water, Wetland, and Cropland exhibited consistently high F1-scores, while performance in low-quality imagery remained coherent despite substantial sensor differences. These findings demonstrate that the algorithm maintains both spatial fidelity and temporal consistency across heterogeneous historical imagery.
*Conclusions* This study shows using GeoAI/ML tools along with multiple sources of data can effectively produce consistent high-resolution multi-decade land-cover maps based on NAIP imagery. The

J. Liu · C. Wang · Z. Yan · Q. Zhang · C. Song (✉)
Department of Geography and Environment, University of North Carolina at Chapel Hill, Chapel Hill, NC 27599, USA
e-mail: csong@email.unc.edu

J. Liu · Y. Dai
Department of City and Regional Planning, University of North Carolina at Chapel Hill, Chapel Hill, NC 27599, USA

X. Tang
Department of Geography, Hong Kong Baptist University, Hong Kong SAR, China

C. Wang
Department of Earth, Marine and Environmental Sciences, University of North Carolina at Chapel Hill, Chapel Hill, NC 27599, USA

approach developed in this study provides a scalable solution for generating high-resolution time series land-cover maps across the conterminous USA where NAIP imagery is available, supporting long-term land-change analyses and landscape-level planning.

## Introduction

Land cover information is fundamental for monitoring environmental and ecological processes. Land cover classification based on remote sensing imagery can provide a foundational dataset for a wide range of applications, including climate modeling (Wulder et al. 2018), biodiversity assessment (Pettorelli et al. 2014) and landscape change analysis (Pielke et al. 2011). Over the past decades, numerous studies based on medium-resolution imagery, such as Landsat (Wulder et al. 2022), have produced long-term land cover products, such as the National Land Cover Database (NLCD), providing critical support for the continuous monitoring of terrestrial processes (Hansen and Loveland 2012). However, sub-meter aerial imagery, such as that collected by the National Agriculture Imagery Program (NAIP), captures much finer boundaries and spatial structures of complex urban and rural surfaces, offering irreplaceable value for accurately detecting and quantifying long-term changes (Nagel and Yuan 2016; Hussain and Shan 2016; Coleman et al. 2020). Despite such an advantage, existing land cover datasets rarely possess such fine spatial details with long-term temporal consistency. Therefore, constructing a land cover dataset that combines high spatial resolution with long-term temporal consistency is essential for accurately tracing historical dynamics, evaluating policy outcomes (Hansen et al. 2013), and modeling future scenarios (Hoffmann et al. 2023).

Obtaining long-term and consistent high-resolution land cover products faces a core challenge—the temporal heterogeneity of archival imagery (Van Den Broeck et al. 2022). This heterogeneity arises because acquisition standards and imaging technologies in NAIP have continuously evolved over the past two decades, leading to notable inter-annual variations in spatial resolution, spectral composition (e.g., the transition from three-band RGB to four-band RGB–NIR), sensor characteristics, and radiometric consistency (Prior et al. 2022; Bhatt and Maclean 2023). These cross-temporal inconsistencies violate the assumption of data homogeneity (Du et al. 2002; Canty and Nielsen 2008; Prior et al. 2022), making it difficult for a single, unified modeling strategy to maintain stable performance across a multi-decadal time span (Tuia et al. 2016; Luo and Ji 2022).

Early attempts to generate multi-temporal land-cover maps primarily followed two steps: (1) training a classifier based on training samples and (2) classification of images based on radiometric normalization of time-series images to the image where the training samples were collected, i.e., signature extension (Minter 1979; Gray and Song 2013). However, radiometric normalization cannot effectively address nonlinear signature drift that originates from multiple sources, which include the sensor characteristics, atmospheric condition, viewing and illuminating geometries, and phenology, leading to unstable temporal trajectories (Lu and Weng 2007). To address this limitation, the Automatic Adaptive Signature Generalization (AASG) framework was proposed which uses stable pixels as temporal anchors to ensure signature consistency between the training samples and the image to be classified. The AASG has been shown to be effective for classification of long time-series Landsat images (Gray and Song 2013; Dannenberg et al. 2018). However, the requirement of sub-pixel accuracy for geometric registration is far more challenging to achieve for sub-meter NAIP imagery than Landsat imagery, making AASG (a method that relies on radiometric consistency) unsuitable for direct application to NAIP imagery.

In recent years, deep learning–based pretrained foundation models have shown outstanding abilities in feature learning and knowledge transfer for remote sensing image interpretation, substantially reducing dependence on task-specific training samples under limited-annotation conditions (Wang et al. 2022; Yao et al. 2023; Huo et al. 2025). Existing studies have widely explored zero-shot and few-shot domain transfer paradigms to address data inconsistencies caused by differences in spatial regions or sensor types (Xu et al. 2022, 2025). However, most of these studies focused on horizontal model transfer between high-resolution images of consistent quality (Tong et al.

2020; Li et al. 2023), while temporal heterogeneity across multi-year archives remains largely unaddressed (Tuia et al. 2016; Luo and Ji 2022). Therefore, effectively leveraging advanced models to generate land cover products that are both temporally consistent and spatially detailed remains a critical scientific problem yet to be solved (Tang et al. 2025).

Existing zero-shot and few-shot domain transfer methods have mainly targeted at high-resolution imagery with consistent quality (Xu et al. 2022). These approaches typically fine-tune pretrained models by combining pseudo-label and sample selection with patch-based classification and object- or hierarchy-based segmentation (Hossain and Chen 2019a; Tong et al. 2020; Li et al. 2023). Their goal is to balance class identification with boundary precision across sensors or regions. However, for heterogeneous aerial archives such as NAIP, imagery acquired within the same year can still show pronounced domain discrepancies across regions. Consequently, few-shot fine-tuning and pseudo-labeling cannot ensure stable model transfer under such conditions. Moreover, current studies remain limited to model transfer itself and lack an operational workflow for producing long-term, high-quality land cover maps across diverse temporal and quality conditions (Luo and Ji 2022; Tang et al. 2025).

To address this challenge, this study proposes and validates a dual-track adaptive workflow designed to generate temporally consistent and reliable 1-m resolution land-cover maps from the highly heterogeneous NAIP archive, using GeoAI/ML tools and existing land-cover products (Maxwell et al. 2019). The core idea of this framework is to apply differentiated processing strategies according to image quality. For the high-quality modern epoch (2009–2017), a pretrained deep learning model is used for robust initial classification (ArcGIS Online 2025) which has been extensively trained and validated on large volumes of NAIP imagery and is seamlessly integrated within the ArcGIS Pro environment. This ensures methodological stability, operational reproducibility, and accessibility for long-term land-cover production workflows. And subsequently, the visual foundation model, Segment Anything Model (SAM) (Kirillov et al. 2023), is incorporated to refine class boundaries—particularly for the delimitation (Ma et al. 2024; Janowski and Wróblewski 2024). For the low-quality early epoch (2004–2008), a data-fusion and back-casting pipeline is developed, using high-quality classifications and the NLCD as a spatiotemporal bridge. This approach enables the reliable reconstruction of historical land-cover information without additional labeling or model retraining for those years. Overall, this study developed an innovative GeoAI/ML approach that enables temporally heterogeneous high-resolution imagery to produce consistent long-term high-resolution land cover maps.

## Materials and methods

### Study area

This study centers on six counties in two Eastern U.S. states: Sampson, Wayne, and Wilson counties in North Carolina (NC), and Blair, Cambria, and Huntingdon counties in Pennsylvania (PA) (Fig. 1). These areas were selected to support an NIH project (R21-MH140292-01) that aims at evaluating long-term exposure of children to green space on their mental health. The historical NAIP imagery archives systematically capture the program's technological evolution ranging from early lower-quality imagery (e.g., 2004) to recent high-quality acquisitions (e.g., 2017), and encompassing variations in spectral composition, sensor characteristics, and radiometric consistency (Homer et al. 2020; Schroeder et al. 2022). This makes them ideal case studies for examining the core challenge addressed in this research—the pronounced temporal heterogeneity across more than a decade of high-resolution imagery (Maxwell et al. 2019; Prior et al. 2022; Bhatt and Maclean 2023).

In addition, the two study regions differ markedly in both physical geography and human–landscape patterns (Omernik and Griffith 2014). The selected NC region, located in the Atlantic Coastal Plain, features predominantly flat terrain characterized by intensive agriculture and more dispersed patterns of recent urban expansion. In contrast, the selected PA region, situated within the Appalachian Ridge and Valley province, exhibits pronounced topographic relief, dominated by extensive deciduous forests and compact post-industrial urban forms. This strong geographical and socio-environmental contrast provides an ideal testbed for evaluating the spatial generalization, robustness, and transferability of the proposed algorithm across regions with diverse spectral signatures, topography, and land use configurations (Maxwell et al. 2019; Homer et al. 2020).

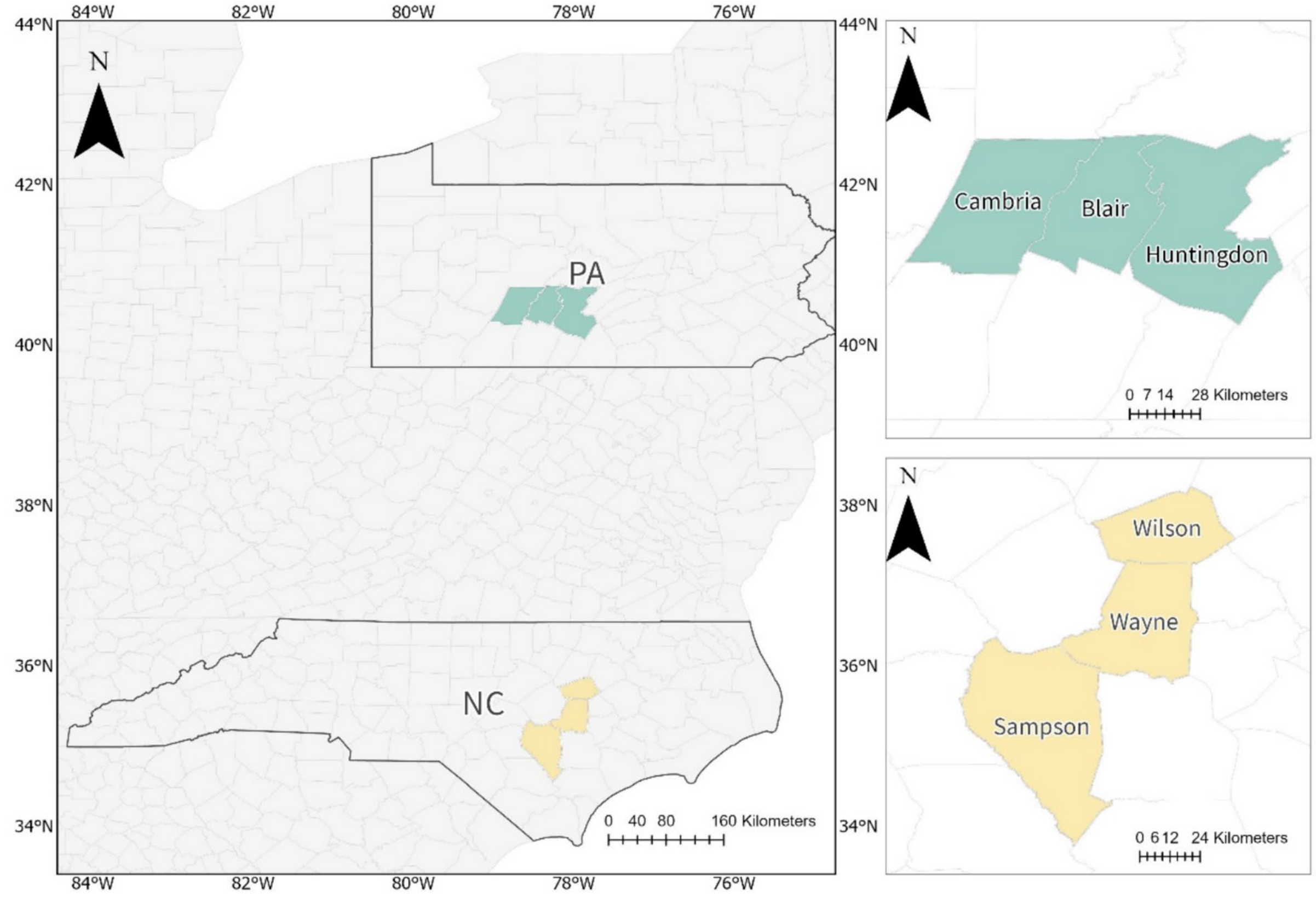

**Fig. 1** Study areas in North Carolina (NC) and Pennsylvania (PA), the United States

## Data

### *NAIP imagery*

This study primarily utilizes the National Agriculture Imagery Program (NAIP) orthoimage archive, which contains high-resolution aerial imagery across the contiguous United States to support agricultural and environmental applications (Schroeder et al. 2022). NAIP imagery collected between 2004 and 2017 was employed in this study. Only RGB bands were used to ensure consistent spectral inputs across the full 2004–2017 archive. Incorporating NIR from later years would incur band inconsistency for the foundation model used in the study.

### *Ancillary datasets*

Two auxiliary datasets at 30 m spatial resolution were incorporated to support and enhance the classification workflow. The first is the National Land Cover Database (NLCD), released by the U.S. Geological Survey (USGS), which provides multi-temporal nationwide land cover information (Yang et al. 2018). In our study, the NLCD served as a key spatiotemporal bridge within the back-casting module, helping to identify potentially changed and stable areas and to supply cross-year reference cues. The second is the Cropland Data Layer (CDL), produced by the U.S. Department of Agriculture (USDA), offering annual crop-type and spatial distribution data (USDA National Agricultural Statistics Service (NASS), 2024). The CDL was incorporated as a raster-based auxiliary layer to supply crop-type priors and seed masks, which guided the refinement of cropland delineation and complemented the missing cropland class in the initial foundation model classification results.

## The dual-track adaptive workflow

To address the pronounced temporal heterogeneity of the NAIP imagery archive, we developed a dual-track adaptive workflow that applies different processing strategies according to image quality. For high-quality years (2009–2017), land-cover mapping is performed using a pretrained foundation model followed by object-level boundary refinement with Segment Anything Model (SAM) to achieve high accuracy (Track 1); For low-quality years (2004–2008), a composite approach integrating NLCD-based spatiotemporal bridging, label back-casting, and object-level optimization is employed to reliably reconstruct historical classifications without additional training samples (Track 2). This step is similar to the concept developed in the AASG (Gray and Song 2013). The overall framework is illustrated in Fig. 2. The NLCD-based propagation strategy in track 2 was adopted solely for early years lacking reliable high-resolution training labels. In contrast, track 1 directly leverages a pretrained foundation model trained on extensive NAIP-aligned reference data, making additional NLCD-based supervision unnecessary.

### *Track 1: Classification of high-quality imagery (2009–2017)*

For high-quality NAIP imagery (2009–2017), we first applied ESRI's pretrained high-resolution land-cover classification foundation model, implemented within the ArcGIS Living Atlas of the World (ArcGIS Online 2025). This model, based on a U-Net architecture with a ResNet-34 backbone, produces a robust baseline 9-class pixel-level prediction (Water, Wetland, Tree Canopy, Shrubland, Low Vegetation, Barren, Structure, Impervious Surface, and Impervious Road) with a reported overall accuracy of approximately 86.5% (ArcGIS Online 2025). This model does not include cropland as a standalone category by design; agricultural areas are generally represented within the broader "Low Vegetation" class. We then employed the SAM for cropland boundary delineation. Pretrained on large-scale visual datasets, SAM provides effective zero-shot generalization for object

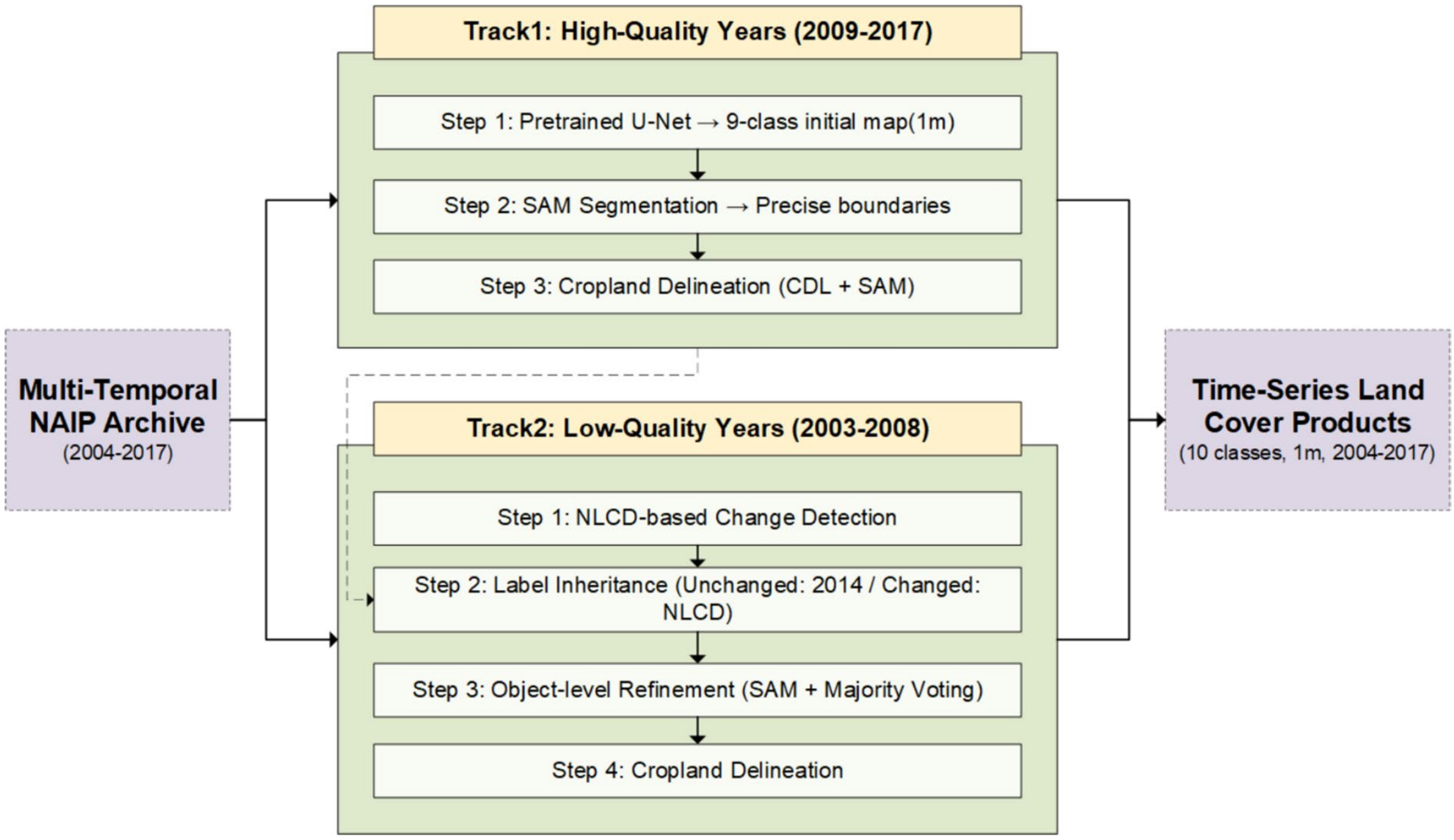

**Fig. 2** Flowchart illustrates the dual-track adaptive workflow for producing temporally consistent and long-term land-cover classifications from heterogeneous NAIP imagery

and boundary delineation in remotely sensed images (Kirillov et al. 2023). Because SAM performs most reliably on compact and spatially contiguous regions, it is particularly suitable for delineating agricultural parcels. Following the implementation settings of SAM for high-resolution remote sensing imagery proposed by Ma et al. (Ma et al. 2024), we applied SAM to the NAIP imagery to produce well-defined feature objects, thereby improving geometric fidelity while preserving the semantic integrity of the pixel-level classification. After generating the land boundaries, we derived the cropland class through a region-growing method using 30 m CDL pixels as seed points, which were expanded only within the 1 m object boundaries generated by SAM. This hybrid approach mitigated the scale mismatch between the CDL and NAIP data, and the resulting 1 m cropland layer was merged with the nine-class base map to produce the complete ten-class land cover product.

*Track 2: Back-casting for low-quality imagery (2004–2008)*

The core challenge in this study is to generate the reliable classification of low-quality images in early years (2004–2008). Early NAIP acquisitions were commonly delivered at 2 m spatial resolution. In addition, acquisition dates, sensor configurations, and radiometric characteristics varied across states and flight cycles, leading to substantial cross-year heterogeneity relative to later standardized 1 m products. To mitigate these issues, we developed a change-detection strategy that leverages the NLCD as a stable inter-annual bridge. We selected 2014 as the high-quality reference year because its NAIP imagery was used for training the ESRI foundation model and it represents a reasonable temporal midpoint within the archive. For each low-quality target year (e.g., 2004), the corresponding 2004 NLCD map was compared with the 2014 reference NLCD map. Areas retaining identical NLCD classes were labeled as unchanged, whereas those with different classes were flagged as changed, producing a 30 m resolution land-cover change mask. This approach implicitly assumes that NLCD stability at 30 m approximates stability at 1 m resolution. While NLCD stability does not preclude fine-scale within-class modifications (e.g., small structures, minor canopy loss, or urban infill), it provides a pragmatic constraint for maintaining cross-year compositional consistency in the absence of consistent high-resolution historical imagery.

Based on the 30 m change mask generated in the previous step, label propagation was conducted according to two rules: pixels within unchanged areas directly inherited land-cover labels from the high-quality 2014 classification results, whereas pixels within changed areas were first assigned preliminary labels from the target year's NLCD map (e.g., the 2004 NLCD product), meaning no supervised classifier was involved at this stage. Because the preliminary labels assigned to the changed areas were derived from the 30 m NLCD map, their boundaries may be coarse and spatially ambiguous. To refine class boundaries within NLCD-changed areas, we applied SAM to the original NAIP imagery to generate geometrically coherent object boundaries. Only SAM-derived objects intersecting the change mask were retained. For each retained object, a single land-cover class was assigned according to the dominant NLCD-derived pixel class of the target year within that object. All 1 m pixels inside the object then inherited this assigned label. This object-level relabeling procedure replaced the coarse 30 m pixel-based assignments within changed areas, resulting in a geometrically coherent 1 m land-cover map, while areas not in change mask largely preserved their inherited labels from the 2014 reference classification, except where minor boundary adjustments occurred due to object-level refinement. To complete the historical land-cover maps, agricultural land was delineated using the same region-growing framework as in Track 1. In years where CDL coverage was available, CDL cropland pixels were used as primary seeds; in years or counties without CDL coverage (e.g., 2004 in the eastern study region), harmonized NLCD cropland pixels were used as fallback seeds. In both cases, a $<1$ ha minimum seed-area threshold and SAM-constrained object boundaries were applied to ensure methodological consistency across years. Importantly, the objective of this framework is to generate a temporally consistent high-resolution land-cover dataset suitable for landscape-scale exposure assessment and regional compositional analysis, rather than to reconstruct parcel-level historical detail for fine-scale change detection.

### Class harmonization and crosswalk

NLCD developed subclasses (21–24) were merged into a single "Developed" category during change-mask generation to prevent minor intensity transitions from being misclassified as change. Because NLCD does not distinguish impervious subclasses (e.g., structures, roads, and other impervious surfaces), these categories were treated as a single "Developed" class at the 30 m level during change-mask generation. The finer subclass distinctions in the final 1 m maps originate from the HRLC foundation model (Track 1) and are preserved through label inheritance in NLCD-unchanged areas. In NLCD-changed areas, only harmonized class-level labels are assigned based on the 30 m NLCD data, and we do not attempt to infer historical subclass transitions unsupported by the NLCD legend. Other categories (e.g., barren, shrubland, grassland, cropland) follow direct one-to-one mappings between HRLC and NLCD without ambiguity.

A pixel was defined as "unchanged" only when the harmonized NLCD class was identical between the target year and 2014 at the 30 m pixel level.

### Accuracy assessment

To systematically evaluate the performance of our dual-track workflow across the full spectrum of NAIP data quality, we conducted a detailed accuracy assessment for three representative years: 2004, 2014, and 2017. These years were carefully selected to capture distinct conditions and challenges within the time series: 2004 represents the earliest and lowest-quality imagery, providing a rigorous test of the Track 2 back-casting method; 2014 serves as a critical benchmark year, functioning as the high-quality reference for workflow; and 2017 represents the most recent and highest-quality imagery, enabling the evaluation of Track 1 performance under near-ideal conditions.

For each of these three years, an independent validation dataset was generated through a multi-stage stratified sampling design (Olofsson et al. 2013, 2014). First, fifty geocoded household addresses were randomly selected from a list of approximately 1,000 residential addresses in our cohort across the six counties and used as sampling centers. A 1 km radius buffer was created around each selected address. Within these buffers, ground-truth land-cover classes were determined through detailed visual interpretation of the corresponding NAIP imagery. For each land-cover class present in a buffer, two candidate validation points were randomly generated. All candidate points from the 50 buffers were then pooled, and a final validation set was constructed by stratified random selection of 50 validation points per land-cover class. Using this validation dataset, we generated confusion matrices for each of the three classification maps (2004, 2014, and 2017) and computed Overall Accuracy (OA), Producer's Accuracy (PA), User's Accuracy (UA), and F1-score. The Kappa coefficient was also reported for completeness, although it is known to have limitations in land-cover accuracy assessment (Pontius and Millones 2011).

### Implementation details

All geospatial processing and deep learning inferences were conducted within the ArcGIS Pro (Version 3.4) software environment. The 9-class baseline classification for Track 1 was generated using ESRI's pretrained high-resolution land-cover classification foundation model (U-Net/ResNet-34 backbone). Key inference parameters were set as follows: padding = 128, batch_size = 8, tile_size = 512, predict_background = True, detailed_classes = True, and test_time_augmentation = True. These parameter values follow the recommended configuration for ESRI's high-resolution land-cover classification foundation model (Item ID: a10f46a8071a4318bcc085dae26d7ee4; accessed May 2025) (ArcGIS Online 2025).

Object segmentation with Meta AI's Segment Anything Model (SAM) was employed locally using Python (3.9) in a Jupyter Notebook. We utilized the ViT-H model checkpoint (sam_vit_h_4b8939.pth) with the SamAutomaticMaskGenerator and the following parameters: pred_iou_thresh = 0.96, points_per_side = 16, points_per_batch = 32, crop_nms_thresh = 0.5, and box_nms_thresh = 0.5. The imagery was processed in patches using a patch_size of 512 and a stride of 256. These settings are consistent with commonly used configurations in SAM applications on high-resolution aerial imagery and were found to produce stable and geometrically precise boundaries (Ma et al. 2024). We conducted limited exploratory adjustments to key SAM parameters within reasonable ranges; however, these adjustments did not yield consistent improvements in object boundary quality or downstream class-level accuracy. We therefore retained the parameter settings reported in Ma et al. (2024).

**Table 1** Crosswalk between HRLC, NLCD, and the final 10-class legend

| Final 10-class legend | HRLC class | NLCD class codes |
|---|---|---|
| Water | Water | 11 |
| Wetland | Wetland | 90, 95 |
| Tree canopy | Tree | 41, 42, 43 |
| Shrubland | Shrub | 52 |
| Grassland | Low Veg | 71, 72, 81 |
| Barren | Barren | 31 |
| Structures | Structures | 21, 22,23,24 |
| Impervious surfaces | Impervious Surfaces | 21, 22,23,24 |
| Impervious roads | Impervious Road | 21, 22,23,24 |
| Cropland | | 82 |

**Table 2** Overall accuracy metrics for 2004, 2014, and 2017

| Year | Area-weighted OA | 95% CI | Kappa | Macro F1 |
|---|---|---|---|---|
| 2004 | 0.733 | [0.669, 0.796] | 0.764 | 0.784 |
| 2014 | 0.887 | [0.843, 0.930] | 0.860 | 0.871 |
| 2017 | 0.886 | [0.842, 0.928] | 0.831 | 0.849 |

All computations were performed on a workstation equipped with a 13th-generation Intel i9 CPU, (32 GB) of RAM, and an NVIDIA GeForce RTX 4080 GPU. With this configuration, processing a 4 $km^2$ area with the SAM mask generator took approximately 5 min and 1.5 min with ESRI's pretrained high-resolution land-cover classification foundation model (Table 1).

## Results

### Overall classification accuracy

The overall accuracy metrics for the three representative years are summarized in Table 2. Because validation samples were selected using equal-per-class stratified sampling, map-wide overall accuracy (OA) was estimated using an area-weighted estimator. The area-weighted OA values are 0.733, 0.887, and 0.886 for 2004, 2014, and 2017, respectively, with corresponding 95% confidence intervals of [0.669, 0.796], [0.843, 0.930], and [0.833, 0.918], respectively, with corresponding Kappa coefficients of 0.764, 0.860, and 0.831. The Macro F1-scores exhibited a consistent pattern (0.784, 0.871, and 0.849). The highest accuracy was recorded for 2014. In contrast, performance in 2004 was lower as expected, given the absence of high-quality training data and the dependence on NLCD-based backcasting strategy. Nevertheless, the area-weighted OA of 0.733 remains within an acceptable range for long-term reconstruction under limited reference-data conditions, indicating that the workflow can produce reliable land cover maps across heterogeneous historical imagery.

### Per-class performance

The per-class performance metrics, including User's Accuracy (UA), Producer's Accuracy (PA), and F1-scores, are summarized in Table 3. An inter-annual comparison of these results reveals several consistent and significant patterns. The Structure and Water classes demonstrated exceptionally stable and high performance, consistently maintaining F1-scores above 0.89 across all evaluated years, which indicates the workflow's consistent reliability in identifying built-up features and water bodies. Similarly, the 'Cropland' class, generated using our hybrid approach of CDL-seeding constrained by SAM-delineated boundaries, achieved high F1-scores ranging from 0.842 to 0.891, validating the effectiveness of the strategy. The Tree Canopy and Wetlands classes also showed robust performance, with F1-scores generally exceeding 0.77 across all years. In contrast, the Grassland class exhibited greater spectral ambiguity, resulting in comparatively lower F1-scores that ranged from 0.687 to 0.796. Notably, classes that are often challenging to distinguish, such as Shrubland, Barren, Impervious Surfaces, and Roads, achieved strong results within our framework, with F1-scores for most categories surpassing 0.80 in the higher-quality years of 2014 and 2017.

**Table 3** Per-class accuracy metrics for 2004, 2014, and 2017

| Class | PA-2004 | UA-2004 | F1-2004 | PA-2014 | UA-2014 | F1-2014 | PA-2017 | UA-2017 | F1-2017 |
|---|---|---|---|---|---|---|---|---|---|
| Water | 0.927 | 0.760 | 0.835 | 0.957 | 0.880 | 0.917 | 1.000 | 0.800 | 0.889 |
| Wetland | 0.750 | 0.960 | 0.842 | 0.875 | 0.840 | 0.857 | 0.732 | 0.820 | 0.774 |
| Tree canopy | 0.625 | 0.800 | 0.702 | 0.868 | 0.920 | 0.893 | 0.977 | 0.840 | 0.903 |
| Shrubland | 0.765 | 0.520 | 0.619 | 0.83 | 0.880 | 0.854 | 0.745 | 0.760 | 0.752 |
| Grassland | 0.694 | 0.680 | 0.687 | 0.943 | 0.660 | 0.776 | 0.812 | 0.780 | 0.796 |
| Barren | 0.75 | 0.660 | 0.702 | 0.972 | 0.700 | 0.814 | 0.887 | 0.940 | 0.913 |
| Structure | 0.891 | 0.980 | 0.933 | 0.926 | 1.000 | 0.962 | 0.904 | 0.940 | 0.922 |
| Impervious surface | 0.925 | 0.740 | 0.822 | 0.786 | 0.88 | 0.83 | 0.867 | 0.780 | 0.821 |
| Impervious road | 0.766 | 0.980 | 0.860 | 0.847 | 1.000 | 0.917 | 0.796 | 0.860 | 0.827 |
| Cropland | 0.889 | 0.800 | 0.842 | 0.817 | 0.980 | 0.891 | 0.828 | 0.960 | 0.889 |

## Confusion matrices

To further examine category-level classification errors, Fig. 3 presents the confusion matrices for 2004, 2014, and 2017. Overall, a strong diagonal dominance is evident, indicating high agreement between predicted and reference classes across most land-cover categories. In the high-quality years (2014 and 2017), misclassifications were largely confined to spectrally or structurally similar classes, such as Grassland vs. Cropland and Shrubland vs. Low Vegetation, consistent with the per-class F1 patterns summarized in Table 3. In contrast, the 2004 results display more widespread off-diagonal errors, primarily among vegetation-related categories, reflecting the limitations of the NLCD-based back-casting approach in distinguishing fine-scale vegetation structures within lower-quality imagery. Nevertheless, key land-cover types such as Water, Structure, and Impervious Roads remained accurately identified across all years, underscoring the workflow's robustness and temporal consistency even under heterogeneous image qualities.

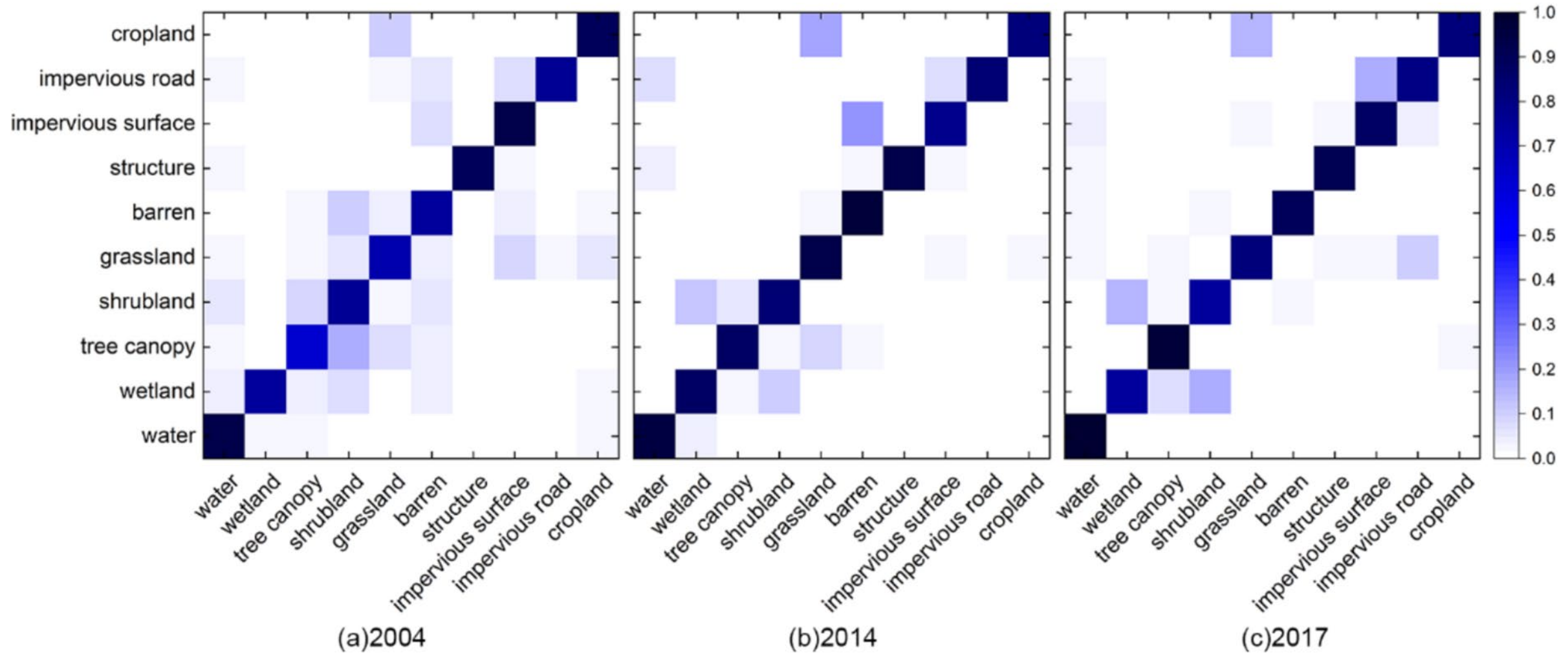

**Fig. 3** Confusion matrices for **a** 2004, **b** 2014, and **c** 2017 classification results. The diagonal dominance indicates strong classification accuracy across most classes, while off-diagonal entries highlight confusion between spectrally similar categories such as grassland and cropland

Ablation analysis

To evaluate the contribution of SAM-constrained cropland completion in Track 1, we compared the proposed method with a CDL-only baseline. The baseline directly used the 30 m CDL cropland layer, whereas the proposed approach applied region-growing constrained by SAM-derived object boundaries.

To isolate the impact of the SAM constraint, evaluation was restricted to areas where the two outputs differed. A 5 m buffer was applied to the difference mask to capture boundary-adjacent regions potentially affected by object refinement. Within this modified zone, cropland boundaries were delineated from each output and a 5 m buffer was constructed to define candidate edge regions. From these areas, 100 validation points were sampled (60 boundary and 40 interior points). Overall accuracy and edge-specific accuracy were then computed to distinguish geometric boundary effects from interior class consistency. The proposed method substantially outperformed the CDL baseline, achieving higher overall accuracy (0.94 vs. 0.64) and improved edge accuracy (0.96 vs. 0.75) (Table 4).

To assess the effect of SAM-based object refinement in Track 2(step 3), we compared the track 2 output with SAM refinement against an otherwise identical version of without SAM refinement step. All other components, including the NLCD-derived change mask and the label assignment procedure are kept in same.

A different mask was generated and expanded using a 5 m dilation to define the modified zone to capture the pixel potentially affect by the boundary refinement. Because several land cover class were sparsely represented within modified zone, the original nine classes were consolidated into three dominant structural categories to ensure statistically stable and interpretable accuracy. Water and wetland were excluded because there were sparsely distributed within the modified zone. Class boundaries were delineated and a 5 m buffer was used to define candidate edge regions. Validation points were sampled from both boundary and interior areas, and overall accuracy and edge-specific accuracy were computed within the modified zone.

Table 5 summarizes the overall comparison between the NoSAM and Smoothed outputs within the modified zone. The Smoothed version achieves higher overall accuracy (0.847) than NoSAM (0.473). Edge accuracy also increases from 0.596 (NoSAM) to 0.875 (Smoothed). Macro F1 improves from 0.502 to 0.853, and balanced accuracy increases from 0.531 to 0.862.

Table 6 presents class-specific accuracy results within the modified zone. For Woody Vegetation, overall accuracy increases from 0.340 (NoSAM) to 0.839 (Smoothed), and edge accuracy increases from 0.361 to 0.849. For Low Vegetation, overall accuracy increases from 0.309 to 0.800, and edge accuracy increases from 0.451 to 0.820. For Impervious Surface, overall accuracy increases from 0.857 to 0.921, and edge accuracy increases from 0.921 to 0.946. Interior accuracy values for each class are also reported in Table 6. Interior impervious samples

**Table 4** Track 1 ablation results (CDL-only vs. SAM-constrained cropland completion)

| Model | OA | Edge accuracy | N_overall | N_boundary |
|---|---|---|---|---|
| Proposed method | 0.94 | 0.96 | 100 | 60 |
| CDL | 0.64 | 0.75 | 100 | 60 |

**Table 5** Track 2 ablation: Overall performance comparison within modified zone

| Subset | Model | OA | Macro F1 | Balanced accuracy |
|---|---|---|---|---|
| Overall | NoSAM | 0.473 | 0.502 | 0.531 |
| Overall | Smoothed | 0.847 | 0.853 | 0.862 |
| Edge | NoSAM | 0.596 | 0.577 | 0.602 |
| Edge | Smoothed | 0.875 | 0.872 | 0.876 |

**Table 6** Track 2 ablation: Class-specific accuracy comparison within modified zone

| Subset | Model | Woody vegetation | Low vegetation | Impervious surface |
|---|---|---|---|---|
| Overall | NoSAM | 0.340 | 0.309 | 0.857 |
| Overall | Smoothed | 0.839 | 0.800 | 0.921 |
| Edge | NoSAM | 0.361 | 0.451 | 0.921 |
| Edge | Smoothed | 0.849 | 0.820 | 0.946 |
| Interior | NoSAM | 0.311 | 0.051 | – |
| Interior | Smoothed | 0.824 | 0.769 | – |

were insufficient within the modified zone for stable estimation.

Visualizing spatial–temporal consistency

Figures 4 and 5 illustrate examples of the land-cover classification results and corresponding NAIP imagery from two representative areas: a dispersed residential landscape in North Carolina (NC) and a compact urban area in Pennsylvania (PA). The visual comparisons highlight substantial variations in radiometric tone, sharpness, and feature clarity across different years and regions—differences primarily attributed to sensor upgrades, acquisition timing, and atmospheric conditions. Despite these pronounced discrepancies, the proposed dual-track workflow consistently maintained high spatial accuracy and temporal coherence. In both regions, cropland boundaries closely align with their true extents, demonstrating the effectiveness of integrating 30 m CDL-derived seed information with 1 m SAM-delineated object boundaries. Furthermore, the workflow successfully captured realistic land-cover transitions over time. In the NC example (Fig. 4), cropland in the upper section of the 2004 image was converted into residential land by 2014 and 2017; in the PA example (Fig. 5), a forested area visible in 2004 had transitioned into residential and built-up zones by 2014, with additional structures appearing by 2017.

## Discussion

Interpretation of overall performance

The 2014 results (OA = 0.887) are on par with the publicly reported accuracy of the ESRI High-Resolution Land Cover (HRLC) model (OA = 0.87), indicating that the workflow—while introducing an additional Cropland category—maintained overall performance comparable to the foundation model nine-class baseline (ArcGIS Online 2025). The slightly lower accuracy observed in 2017 (OA = 0.886) reflects seasonal and phenological differences among acquisition dates, which can affect land cover class separability even within high-quality imagery. For our six study counties, the 2014 NAIP scenes were acquired mainly in May and August, while the 2017 scenes span May, June, September, and October. This wider temporal spread increases variability in vegetation greenness and illumination, reducing land-cover class separability even within high-quality imagery.

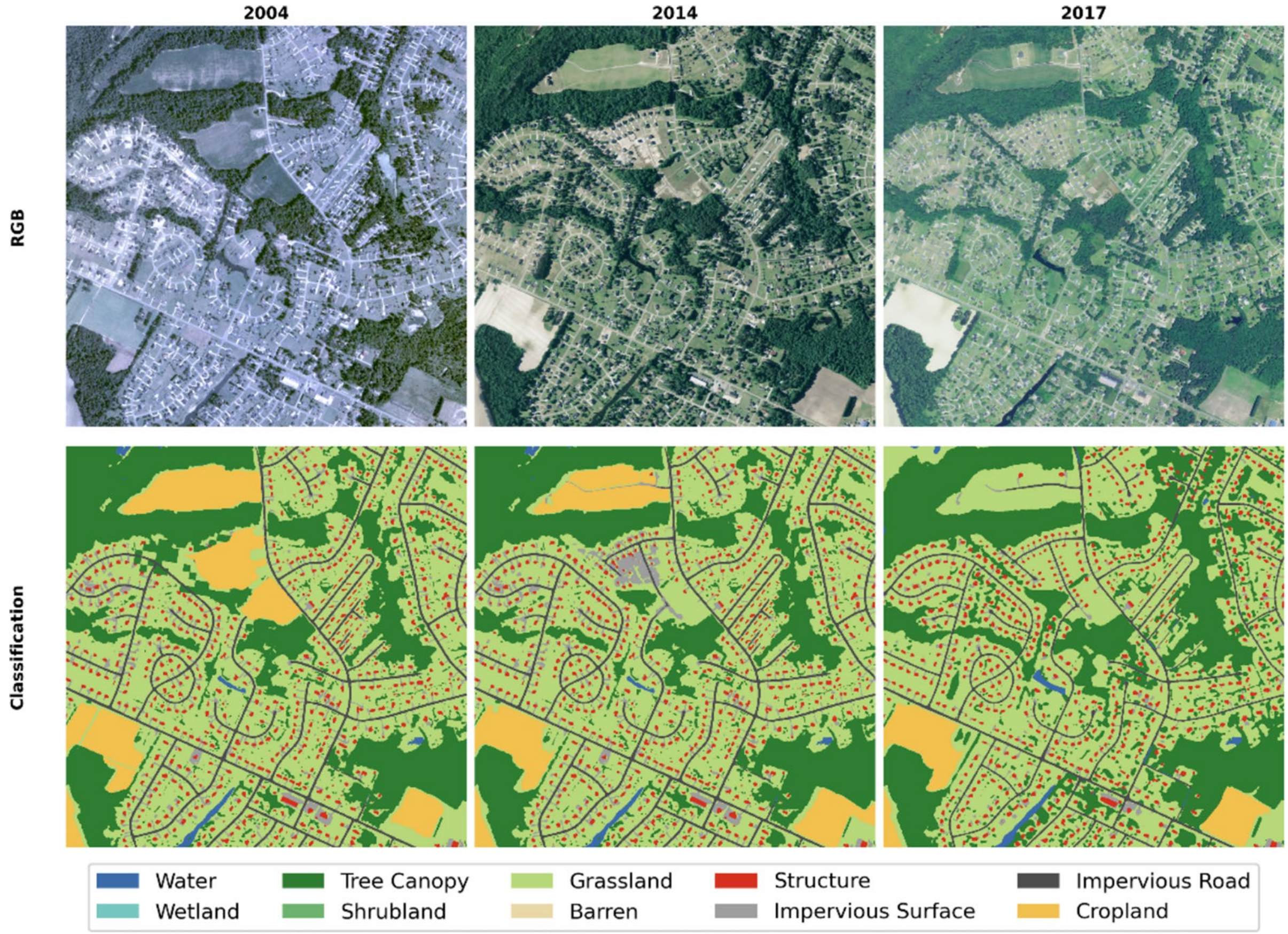

**Fig. 4** A site in North Carolina (NC) featuring low-density residential development

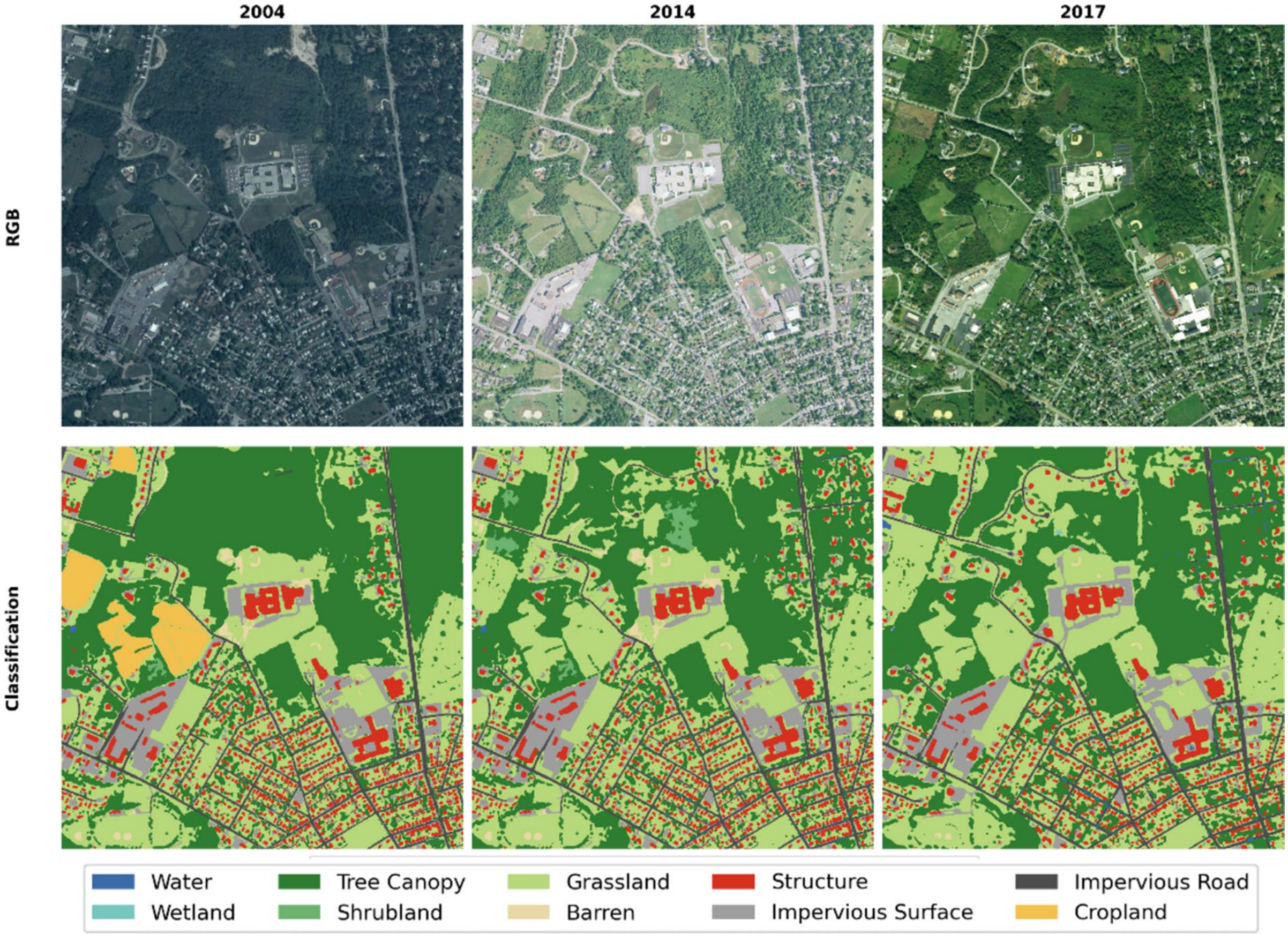

**Fig. 5** A site in Pennsylvania (PA) featuring high-density urban forms and agricultural land

The more pronounced decline in accuracy for 2004 (OA = 0.733) reflects a combination of limitations in early NAIP imagery and uncertainties introduced by the NLCD-based spatiotemporal bridging and label inheritance approach. The primary source of uncertainty lies in propagating 30 m NLCD-derived change information onto the 1 m object-based classification, which can introduce coarse boundaries and miss fine-scale transitions such as small new constructions or localized vegetation changes. To evaluate this effect, we assessed accuracy within NLCD-unchanged cells only (n = 485). The resulting OA (0.792) is comparable to the overall OA (0.733), suggesting that potential within-class changes in NLCD-unchanged areas do not substantially affect class-level consistency in the reconstructed 1 m maps. On the other hand, the lower radiometric and spatial quality of early NAIP imagery may also degrade SAM performance, leading to less accurate object boundaries in these years. This behavior was most evident along forest–impervious interfaces in heterogeneous residential landscapes. In these contexts, SAM occasionally under-segmented narrow, elongated forest strips (e.g., tree lines) embedded within developed areas, resulting in localized boundary simplification along forest–impervious transitions. Given these constraints, an OA of 0.79 for 2004 represents an acceptable outcome and reflects the expected performance range for a back-casting workflow that relies solely on coarse ancillary products. The lower performance in early years also underscores the rationale of the dual-track framework. While the NLCD-based propagation strategy is necessary under limited reference conditions, it is not applied to high-quality imagery in Track 1, where the pretrained foundation model already provides reliable supervision. The distinction between the two tracks reflects differences in data quality and label availability. This result indicates that the workflow remains effective even under minimal reference-data conditions.

### Comparison with ESRI HRLC and interpretation

In the recent years (2014 and 2017), the overall accuracies achieved by our workflow are comparable to those reported for the ESRI High-Resolution Land Cover (HRLC) foundation model (ArcGIS Online 2025), while delivering equivalent or higher F1-scores for several key classes, including Shrubland, Barren, Impervious Surfaces, Impervious Roads, Wetlands, and Structure. Two main factors may contribute to these differences. The first factor is the evaluation domain differences. The ESRI-reported metrics represent nationwide aggregates, whereas our study focuses on six counties in North Carolina and Pennsylvania, which are not only more homogenous but also geographically proximate to the Chesapeake Bay watershed, an area where the HRLC model itself performs comparatively well (ArcGIS Online 2025), thereby benefiting from the proximity to the high performance area. The second factor is the divergent sampling strategies. Our validation employed stratified random sampling within residential address buffers, ensuring representation of both feature interiors and edges. This approach introduces a more realistic proportion of boundary-affected samples than systematic grid sampling, which tends to underestimate errors near class transitions and may inflate PA and UA values. Therefore, our evaluation is conservative rather than optimistic, especially for classes with complex edges.

### The differentiated role of SAM: class completion and boundary refinement

The role of the Segment Anything Model (SAM) differed significantly between the two tracks of our workflow. For recent years (2014, 2017), SAM's primary function was class completion. In conjunction with CDL seeds, it constrained a region-growing algorithm to add the 'Cropland' class, which was absent from the HRLC foundation model, thereby producing a clear, well-defined agricultural layer at 1-m resolution. For other existing categories such as Impervious Surface and Impervious Road, the HRLC foundation model already demonstrated stable semantic predictions and coherent boundaries, as reflected in the per-class accuracy results. Therefore, in Track 1, SAM was applied selectively to enable semantic completion of cropland rather than to globally reassign all classes. In contrast, for the back-casting year (2004), SAM played a more critical role in boundary refinement. To counteract the blocky artifacts introduced by the 30 m NLCD change mask, we selectively retained only the SAM objects that intersected with the change mask. By assigning each SAM object the dominant NLCD-derived class within its boundary, we transformed the coarse, pixelated results into a map with smoother, more realistic object boundaries, partially mitigating the 30 m to 1 m scale mismatch. In essence, SAM's role transitioned from "class completion" in more recent years to "boundary refinement" in earlier years, with both functions serving the overarching goal of inter-annual consistency. Similar object-constrained refinement has been demonstrated in SAM-assisted segmentation of VHR remote sensing imagery, but its differentiated role across temporal tracks has not been discussed in prior studies (Ma et al. 2024). The ablation analyses further clarify the differentiated role of SAM in the two tracks. In both Track 1 and Track 2, incorporating SAM-derived object constraints led to clear accuracy gains, with improvements most evident along class boundaries. In Track 1, SAM primarily strengthens cropland delineation by enforcing coherent parcel geometry, whereas in Track 2 it mitigates the blocky artifacts introduced by pixel-based label propagation. These findings suggest that SAM's principal contribution lies in stabilizing geometric structure and boundary consistency.

### Class-level mechanisms: cropland expansion and statistical bias in grassland

The class-level mechanisms highlight the methodological trade-offs of our workflow. The new Cropland class, introduced via a conservative "CDL (or NLCD fallback) seed + SAM boundary" approach, achieved high F1-scores (0.842–0.891) and produced morphologically coherent polygons with crisp boundaries. This strategy, however, induced an explainable fidelity-omission trade-off in the Grassland class. Excluding small (< 1 ha) cropland patches to ensure seed purity meant that genuine small cropland parcels were often omitted and subsequently absorbed by the Grassland class. This created a stable pattern, corroborated by the confusion matrices and per-class metrics, where Grassland's User's Accuracy was consistently and significantly lower than its Producer's Accuracy—a tunable statistical bias resulting from a methodological choice, rather than a failure of the classification model itself. This limitation is not inherent to the framework but arises from a conservative threshold selection. In future applications, adaptive seed sizes or object-level probability filtering could be incorporated to retain small cropland patches while minimizing false positives. Thus, the observed bias reflects a controllable design decision rather than an irreversible weakness of the workflow. Importantly, this behavior was consistent across years, including 2004 when NLCD cropland was used as a fallback due to CDL unavailability. We did not observe disproportionate degradation in cropland F1-score or abnormal shifts in class-level area proportions in 2004 relative to later years, suggesting that the fallback strategy did not introduce additional systematic bias beyond the intended conservative seed filtering.

### Justification for track 2 and comparison with AI domain adaptation paradigms

The algorithm developed in this study, particularly the Track 2 back-casting method, is fundamentally distinct from end-to-end AI domain adaptation (DA) methods. The development of this algorithm was motivated by an initial assessment in which conventional DA paradigms failed to address the extreme temporal and sensor heterogeneity present in the NAIP archive. We evaluated the two primary AI-based DA paradigms, model transfer and image-to-image translation.

For model transfer, this approach was found to be not viable for two reasons. First, it requires manually labeled data from the target domain (e.g., 2004). However, early NAIP imagery exhibits strong inter-county radiometric variability and acquisition inconsistencies, making it impossible to construct a representative training dataset across a large geographic area, such as that in this study. To avoid manual labeling, we attempted to bypass this by fine-tuning with synthetic data (i.e., down sampling 2014 imagery and applying histogram matching to mimic 2004). However, fine-tuned models produced unstable predictions and severe class biases (e.g., F1-score for Water = 0.14), characterized by a massive over-prediction of spectrally ambiguous classes.

Image-to-Image Translation involves translating the target domain (2004) into the source domain (2014) using a model of Cycle GAN which also did not work well. The generated images suffered from significant visual artifacts, failed to reconstruct valid textures, and were unusable for classification. This qualitative failure was confirmed by quantitatively poor results, with high Fréchet Inception Distance (FID) scores (A: 99.86, B: 111.60).

In contrast to these direct AI adaptation approaches, our Track 2 strategically integrates multiple complementary data sources: the temporally consistent NLCD product, high-quality reference year classifications for stable areas, and SAM-derived boundaries. This integration strategy effectively counters the temporal heterogeneity problem by decomposing it into manageable components, achieving balanced historical reconstruction that direct AI methods failed to deliver.

### Comparison with existing domain-adaptation approaches

Recent cross-domain approaches in high-resolution remote sensing mainly include model transfer or fine-tuning, image-to-image translation, and weakly supervised refinement methods that project low-resolution labels onto high-resolution imagery. However, these paradigms have clear limitations when applied to NAIP's multi-decade, quality-heterogeneous archive. For example, Previous work (Tong et al. 2020) reported that zero-shot cross-sensor transfer in a 15-class setting yielded overall accuracies as low as 50.12% (JL-1(1)) and up to 80.97% (ZY-3(1)), with pseudo-label fine-tuning only partially recovering performance to 52.86% and 82.50%. This fluctuations reflect the unstable performance when applied on different domains。End-to-end feature-alignment frameworks such as Domain Adaptation for Cross-Spatio-Temporal classification (Luo and Ji 2022) achieve state-of-the-art performance on multi-scene and multi-platform benchmarks, substantially improving mean Intersection over Union (mIoU) relative to conventional domain adaptation. Nevertheless, their training cost and architectural complexity pose scalability challenges for multi-decade NAIP time series, where heterogeneous imaging conditions—even across counties within early years。Weakly supervised low resolution to high resolution label-refinement approaches (Tang et al. 2025) improves segmentation quality by 5–15 percentage points by using low-resolution supervision to guide high-resolution learning. Yet, the inherent spatial coarseness of low-resolution labels prevents them from capturing the fine object boundaries and small structures characteristic of 1 m NAIP imagery, such as building footprints or small agricultural parcels.

Compared with the approaches above, our back-casting workflow requires no model-side adaptation. Instead, it uses the high-resolution classification map of the reference year as a geometric and semantic anchor, leverages the temporal consistency of NLCD to identify changed areas, and employs SAM-derived object boundaries to eliminate errors caused by scale discrepancies. This procedure preserves 1-m geometric detail while ensuring semantic transferability. The mechanism avoids the information loss inherent in low-resolution supervision and bypasses the dependence of model transfer and image translation methods on target-domain data, enabling stable and production-ready outputs across NAIP's long-term and quality-heterogeneous archive.

### Future directions

Future research will evaluate the framework in western and semi-arid regions of the United States, where landscape structure, vegetation phenology, and development patterns differ substantially from those in the eastern study domain. Testing the workflow under drier climatic conditions, sparser vegetation cover, and distinct land-use mosaics would provide further insight into its cross-regional transferability and potential adaptation needs.

Beyond regional transferability within the U.S., the dual-track approach could also be extended to other high-resolution multi-temporal datasets. While NAIP provides comprehensive coverage for the United States, similar strategies could be adapted for European aerial photography programs such as Copernicus VHR, EIM, DOP (Germany), or APGB (UK), which also exhibit temporal inconsistencies and sensor upgrades. Applying the workflow to these diverse datasets would further test its robustness across different acquisition standards and landscape configurations.

## Conclusion

This study developed an algorithm that contains a dual-track adaptive workflow to classify the long-term NAIP image collection. The algorithm achieved temporally consistent and spatially detailed 1 m land-cover mapping without retraining or fine-tuning the foundational model. For recent years, the algorithm expands upon the nine-classes HRLC foundation model by incorporating an additional Cropland category; for early years, it completes historical reconstruction through NLCD-based temporal bridging and SAM-guided object-level refinement, generating unified ten-class high-resolution land-cover maps.

Assessment across three representative years across the time series showed that the overall accuracies for 2014 and 2017 were 0.887 and 0.886 (Kappa 0.860 and 0.831), respectively, which is on par with the public baseline. The back-cast result for 2004 achieved a reasonable overall accuracy of 0.733 (Kappa 0.764). Notably, this achievement required no additional training samples for early low-quality imagery, significantly lowering the cost and technical barriers to long-term land cover monitoring. At the class level, key categories such as Structure, Impervious surfaces (including Roads), Wetlands, Barren, and Shrubland performed comparably or better than the baseline, while the newly added Cropland class achieved consistently high accuracy. These results demonstrate that consistent, boundary-preserving, and temporally comparable high-resolution land cover maps can be generated from NAIP image collection even under severe cross-year image quality heterogeneity. The algorithm developed here can be used widely for producing high-resolution land cover maps in other countries with similar data archives.

Beyond classification performance, the primary contribution of this study lies in enabling the reconstruction of temporally consistent high-resolution land-cover maps under conditions of severe cross-year heterogeneity and limited training data availability. Compared with commonly used 30 m products, the 1 m maps generated here preserve consistent object-level geometry and boundary alignment, which improves the stability of landscape-metric estimation across years. This capability supports temporally explicit landscape-scale analyses of spatial configuration, including assessments of fragmentation patterns surrounding residential environments. Such applications are particularly valuable in studies examining long-term environmental exposure, including analyses of neighborhood landscape context and its association with outcomes such as child mental health. In this sense, the contribution of this work extends to support temporally explicit landscape-ecological research.

**Author contributions** JL: Writing—Original Draft, Writing—Review & Editing, Methodology, Software, Validation, Visualization. XT: Writing—Original Draft, Visualization. CW: Writing—Review & Editing, Methodology. ZY: Writing—Review & Editing, Visualization. YD: Writing—Review & Editing. QZ: Writing—Review & Editing. CS: Supervision, Project Conceptualization, Review & Editing,

**Funding** This project was supported by a National Institutes of Health grant (R21-MH140292-01).

**Data availability** The data is available upon reasonable request.

**Declarations**

**Conflicts of interest** The authors declare no competing interests.